\documentclass[a4paper,fleqn]{cas-dc}

\usepackage[authoryear,longnamesfirst]{natbib}

\begin{document}
\let\WriteBookmarks\relax
\def\floatpagepagefraction{1}
\def\textpagefraction{.001}

\shorttitle{Towards human-level performance on automatic pose estimation of infant spontaneous movements}    

\shortauthors{D. Groos et al.}  

\title [mode = title]{Towards human-level performance on automatic pose estimation of infant spontaneous movements}  



%

\author[1]{Daniel Groos}
\author[2,4]{Lars Adde}
\author[2,5]{Ragnhild Støen}
\author[3]{Heri Ramampiaro}
\author[1]{Espen A. F. Ihlen\corref{cor1}}
\cortext[cor1]{Corresponding author: espen.ihlen@ntnu.no (E.A.F. Ihlen)}

\address[1]{Department of Neuromedicine and Movement Science, Norwegian University of Science and Technology, Trondheim, Norway}
\address[2]{Department of Clinical and Molecular Medicine, Norwegian University of Science and Technology, Trondheim, Norway}
\address[3]{Department of Computer Science, Norwegian University of Science and Technology, Trondheim, Norway}
\address[4]{Clinic of Clinical Services, St. Olavs Hospital, Trondheim University Hospital, Trondheim, Norway}
\address[5]{Department of Neonatology, St. Olavs Hospital, Trondheim University Hospital, Trondheim, Norway}









\begin{abstract}
Assessment of spontaneous movements can predict the long-term developmental disorders in high-risk infants. In order to develop algorithms for automated prediction of later disorders, highly precise localization of segments and joints by infant pose estimation is required. Four types of convolutional neural networks were trained and evaluated on a novel infant pose dataset, covering the large variation in 1 424 videos from a clinical international community. The localization performance of the networks was evaluated as the deviation between the estimated keypoint positions and human expert annotations. The computational efficiency was also assessed to determine the feasibility of the neural networks in clinical practice. The best performing neural network had a similar localization error to the inter-rater spread of human expert annotations, while still operating efficiently. Overall, the results of our study show that pose estimation of infant spontaneous movements has a great potential to support research initiatives on early detection of developmental disorders in children with perinatal brain injuries by quantifying infant movements from video recordings with human-level performance.
\end{abstract}




\maketitle

\section{Introduction}
\label{sec:introduction}

During the first months of life, spontaneous infant movements may indicate later developmental disorders, such as cerebral palsy (CP), Rett syndrome, and autism spectrum disorder~\citep{novak2017early,einspieler2014highlighting,einspieler2005early}. Early identification of infants at high risk for developmental disorders is essential in order to successfully select appropriate follow-up approaches, and is of greatest importance in research to evaluate early interventions~\citep{stoen2017computer}. The expert-based observation of general movements (GMs) from video recordings, known as the general movement assessment (GMA)~\citep{ferrari2004prechtl}, has recently been recommended for clinical use in high-risk infants less than five months of age~\citep{novak2017early}. It is especially the fidgety type of GMs, which typically occur between three and five months post-term age, that have shown to predict normal motor development with high accuracy~\citep{einspieler2016fidgety}. However, GMA is dependent on individual expert-based training and interpretations, requires time for video observation and analysis, and triggers a high demand for skilled observers if implemented in large-scale screening~\citep{stoen2017computer}. As an evolving alternative to observational GMA, computer-based methods for objective and consistent risk-assessment are explored~\citep{adde2010early}. This supports clinicians in diagnostics, ultimately identifying infants in need for early interventions and focused follow-up care.

Computer-based assessment of infant movements aggregates quantitative movement information from video recordings to yield estimates for the risk of later disorders (e.g., CP)~\citep{ihlen2020machine}. Hence, higher level of correctness in the representation of movement kinematics, such as segment positions and joint angles, facilitates optimal risk analysis. Fidgety movements are small movements of moderate speed and variable acceleration, of neck, trunk, and limbs, in all directions~\citep{ferrari2004prechtl}. Automated assessment of such movements requires precise localization of the body parts for proper computer-based risk analysis.

The widespread use of conventional video recordings to capture infant movements has established the need for markerless motion capture, which enables the extraction of movement information in an unobtrusive manner~\citep{rahmati2015weakly}. This provides a low-cost alternative to sensor-based motion capture, which can be performed both at the clinic and at home~\citep{adde2021motion}. Markerless motion capture has the potential to make movement assessments more widely available and promotes worldwide collaboration in analysis of infant movements. Moreover, existing large-scale databases of infant recordings, collected by clinical GMA networks~\citep{stoen2019predictive,orlandi2018detection,ferrari2019motor,morgan2019pooled,kwong2019baby,gima2019evaluation}, can be exploited to yield more accurate computer-based methods for risk assessments.

Convolutional neural networks (ConvNets) have improved the techniques for extracting human movement information from conventional 2D videos~\citep{toshev2014deeppose,newell2016stacked,cao2018openpose}. State-of-the-art markerless motion capture tracks movements automatically through frame-by-frame pose estimation, where the ConvNets predict x and y coordinates of a predefined set of body keypoints, directly from the raw video frames~\citep{andriluka20142d}. However, most existing human pose estimation (HPE) methods are targeted towards adults, which compared to infants, differ in anatomical proportions and distribution of body poses~\citep{sciortino2017estimation}. Employed on infant images, the localization performance drops significantly, with 10\% of the estimated body keypoint positions placed outside a head length distance from the annotated ground truth positions (i.e., 90\% in the $PCK_{h}@1.0$ metric described in Section~\ref{sec:metrics})~\citep{sciortino2017estimation}. From this,~\citet{sciortino2017estimation} conclude that there is a need to tune HPE ConvNets to the task of infant pose estimation. 

Following along these lines,~\citet{chambers2020computer} retrains the openly available OpenPose network~\citep{cao2018openpose} by utilizing a dataset of 9 039 manually annotated infant images. This improves infant pose estimation, reducing the mean error by 60\%~\citep{chambers2020computer}. Despite this advance, a recent study carried out by our group found that OpenPose lacks the sufficient scaling of network depth, network width, and image resolution for optimal pose estimation~\citep{groos2020efficientpose}. Other alternatives to OpenPose, such as DeeperCut~\citep{insafutdinov2016deepercut} used in DeepLabCut~\citep{mathis2018deeplabcut}, posses similar shortcomings as single-scale networks targeted towards multi-person pose estimation. Recent developments in HPE outperform OpenPose and variants by deploying novel multi-scale networks and by maintaining higher spatial resolution~\citep{newell2016stacked,sun2019deep}. OpenPose is also computationally inefficient, which makes it less convenient for real-world applications~\citep{groos2020efficientpose}. ConvNet model scaling addresses this challenge by providing trade-offs in localization performance and computational efficiency across various computational budgets~\citep{groos2020efficientpose}, better serving single-person applications. 

The main objective of the present study is to obtain computationally efficient markerless pose estimation of the spontaneous movements of infants with a localization performance approaching that of human expert annotations. We exploit a large and heterogeneous infant pose dataset covering infant recordings from multiple sites across the world to conduct a comparative analysis of the localization performance and computational efficiency of eight different ConvNet models, including the commonly used OpenPose network. We compare the performance level of the ConvNets with the inter-rater spread of human expert annotations.

\section{Materials and methods}
\label{sec:methods}

In this section, we introduce In-Motion Poses, describe the ConvNet models included in the comparative study, and explain the various performance metrics used to evaluate the ConvNets.

\subsection{In-Motion Poses}
\label{sec:inmotionposes}

We developed a dataset comprising infant images with associated human annotations as the ground truth body keypoint positions. We used a large-scale database of 1 424 recordings of 9-18 weeks post-term old infants to facilitate pose estimation of the spontaneous movements of infants in supine position across various recording setups. The videos were collected between 2001 and 2018 through different research projects on observational GMA, and all the recordings follow the standards for video-based GMA during the fidgety movement's period (i.e., infants wear a diaper or a onesie, are awake, alert, and content, are not disturbed or using pacifier, and are positioned in the center of a mattress or blanket with the whole body visible)~\citep{einspieler2005prechtl}. The resolution of videos varied from 576$\times$720 to 1080$\times$1920. The study was approved by the regional committee for medical and health research ethics in Norway, under reference numbers 2011/1811 and 2017/913 on 14 January 2019 and 9 October 2019, respectively. Written parental consent was obtained before inclusion. 

\begin{figure*}[!t]
\centerline{\includegraphics[width=\textwidth]{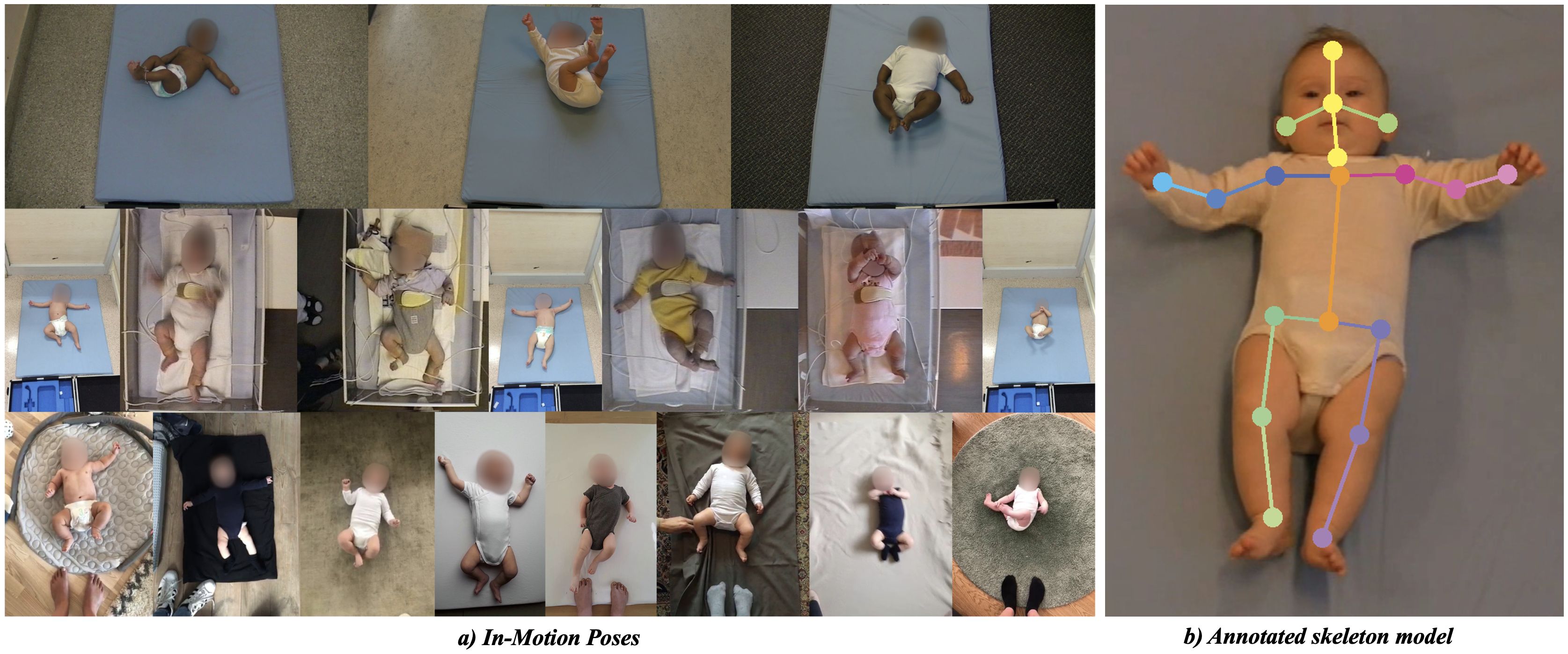}}
\caption{a) A selection of video frames from In-Motion Poses, originating from standardized and less standardized hospital recordings (top and middle row, respectively), and videos captured from home by parents using the In-Motion smartphone application~\citep{adde2021motion} (bottom row). Infant faces are blurred to ensure anonymity. b) The set of 19 body keypoints annotated in the images of In-Motion Poses.}
\label{fig:inmotionposes}
\end{figure*}

From these recordings, we proposed a dataset of 20 000 video frames. The dataset emphasizes the heterogeneity in spontaneous movements by including videos from 12 different sites from seven countries across the globe (i.e., Norway, India, United States, Turkey, Belgium, Denmark, and Great Britain). The videos cover different groups of infants (e.g., typically developing infants, preterm infants, and other high-risk infants enrolled in hospital-based follow-up programs), and are recorded either by clinicians in a hospital setup or by parents using a smartphone application at home~\citep{adde2021motion,stoen2019predictive} (see Fig.~\ref{fig:inmotionposes}a for examples from the dataset). To ensure all video variations were represented, 8 000 (40\%) frames originated from standardized hospital recordings, 8 000 (40\%) from home-based smartphone recordings, and the remaining 4 000 (20\%) from less standardized hospital videos. In each of these three subsets, 80\% of the frames were randomly picked with an equal number of frames from each video. Moreover, to achieve proper variation of infant poses, the remaining 20\% of frames cover infant poses that occur less frequently, and hence might be particularly challenging for an automatic pose estimator. These frames were manually selected from a random pool of 20 000 separate frames (8 000, 8 000, and 4 000 for each subset, respectively), with selection criteria including 1) legs moving towards upper body, 2) overlap of body parts, and 3) crossing of body parts. The resulting total of 20 000 frames were split into training (14 483 (72\%)), validation (1 493 (8\%)), and test sets (4 024 (20\%)) in a common machine learning fashion. To mitigate bias and ensure objective evaluation, all frames of a single infant video were placed into one of these three sets. 

For the ConvNet models to learn from the data in a supervised fashion, and to be able to validate and test the models, the infant images were annotated to produce the ground truth positions. As depicted by Fig.~\ref{fig:inmotionposes}b, 19 distinct body keypoints (i.e., head top, nose, ears, upper neck, shoulders, elbows, wrists, upper chest, right/mid/left pelvis, knees, and ankles) comprised a skeleton model of the infant. The definitions of the body keypoints were agreed upon by a group of human movement scientists and clinical physiotherapists (see Appendix~\ref{sec:keypoints} for a complete overview). Using a separate software tool~\citep{groos2018infant}, 10 human expert annotators (two human movement scientists, two physiotherapists, and six engineers) estimated the x and y coordinates of body keypoints, through manual annotation. All body keypoints were annotated in all images regardless of their type of visibility (i.e., visible or occluded). This resulted in a total of 380 000 human labels (i.e., 19 annotated keypoint positions for each of the 20 000 frames). To measure the consistency between the experts, all annotators estimated the positions of body keypoints in the same sample of 100 randomly selected inter-rater frames. The frames were selected with a similar distribution across recording setups as the full dataset (i.e., 40\% standardized, 40\% home-based, and 20\% less standardized). We computed the inter-rater annotation disagreement in terms of the mean inter-rater spread $H$ of each body keypoint $b$. We calculated the mean distance of an annotation $(x_{b,i,j},y_{b,i,j})$ of an individual expert $j$ of a body keypoint’s position in image $i$, to the average annotation $(\bar{x}_{b,i},\bar{y}_{b,i})$, across the $N$ (i.e., 10) experts for the $S$ (i.e., 100) frames (see~\ref{eq:h}). $H$ was normalized according to the head length of the infant in the image, defined as the distance from the top of the head to the upper neck ($l_{i}$).

\begin{equation}
H_{b}=\frac{1}{N \cdot S} \sum_{i=1}^{S} \sum_{j=1}^{N} \frac{\sqrt{(x_{b,i,j} - \bar{x}_{b,i})^2+(y_{b,i,j} - \bar{y}_{b,i})^2}}{l_{i}}
\label{eq:h}
\end{equation}

\subsection{Comparative analysis}
\label{sec:analysis}

\begin{figure*}[!t]
\centerline{\includegraphics[width=\textwidth]{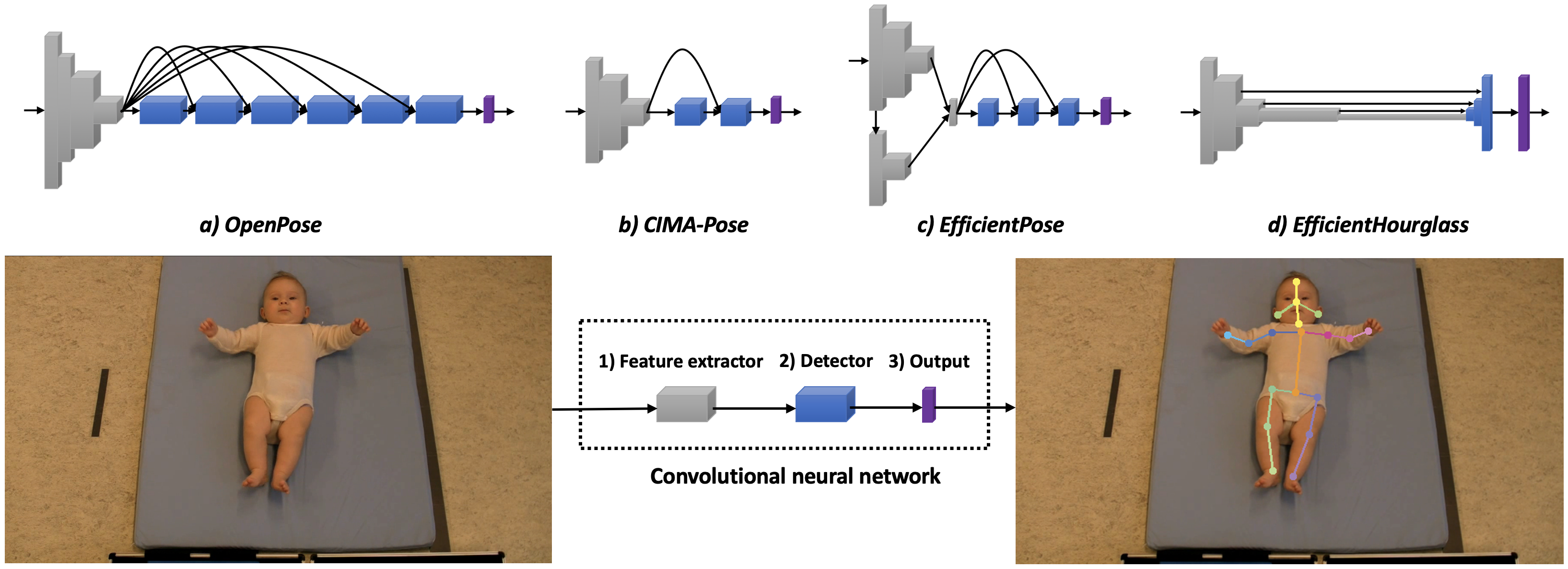}}
\caption{ConvNets address infant pose estimation from video frames in a frame-by-frame manner by 1) extracting image features, 2) determining features relevant for detection, and 3) estimating infant keypoint positions. The height of the ConvNet blocks (i.e., feature extractor, detector, and output) indicates the block’s spatial resolution in relation to the resolution of the input image.}
\label{fig:models}
\end{figure*}

By the use of the aforementioned dataset, we trained and evaluated a selection of ConvNet models for the task of infant pose estimation. First, the ConvNet of the state-of-the-art method for infant pose estimation, the OpenPose network~\citep{cao2018openpose,openpose2021} (see Fig.~\ref{fig:models}a for an architectural overview), was trained to yield baseline performance on In-Motion Poses, while also evaluating the official OpenPose library without fine-tuning\footnote{The latest version of OpenPose (v1.7.0) was used with default settings maintained. Evaluation on In-Motion Poses was performed on the keypoints in the 25-keypoint body model that exist in In-Motion Poses (i.e., all keypoints except head top and upper neck).}~\citep{openpose2021}. Unless otherwise specified, OpenPose refers to OpenPose ConvNet fine-tuned on In-Motion Poses. Second, we trained a more computationally efficient approach inspired by OpenPose, named CIMA-Pose (see Fig.~\ref{fig:models}b), which has displayed promising results on infant pose estimation on videos from standardized clinical setups~\citep{groos2018infant}. CIMA-Pose comprises a ConvNet with low complexity, reflected by 2.4 million parameters compared to 26 million for OpenPose. OpenPose and CIMA-Pose operate on similar image input resolutions of 368$\times$368 pixels\footnote{The raw images in In-Motion Poses were downsampled and zero-padded to square aspect ratio to achieve the input resolution of the ConvNets.}. Third, EfficientPose (Fig.~\ref{fig:models}c) comprises a family of scalable ConvNets demonstrating 57\% improvement in high-precision pose estimation compared to OpenPose, despite significant reduction in computational cost (i.e., FLOPs) and number of parameters~\citep{groos2020efficientpose}. EfficientPose yields five model variants, EfficientPose RT and I-IV, obtained by the use of compound model scaling on input resolution, network width, and network depth. The computational requirements of EfficientPose span from less than one GFLOP to 74 GFLOPs, which is substantially less than the 161 GFLOPs of OpenPose. Fourth and finally, we optimized an EfficientHourglass model with EfficientNet-B4 backbone (i.e., EfficientHourglass B4)~\citep{groos2020approaching}, displayed in Fig.~\ref{fig:models}d. Inspired by the original multi-scale hourglass of~\citet{newell2016stacked}, EfficientHourglass performs parallel processing of image features at different scales, while conserving the level of detail (i.e., resolution) inherent in the input image. With an input resolution of 608$\times$608, EfficientHourglass B4 maintains a resolution of at least 152$\times$152 pixels throughout the stages of the network (i.e., feature extractor, detector, and output), compared to the consistent low resolution of 46$\times$46 pixels in the detector and output of the single-scale OpenPose architecture~\citep{cao2018openpose,groos2020approaching}. For further details of the different ConvNets, the reader is referred to their original papers~\citep{cao2018openpose,groos2020efficientpose,groos2020approaching,groos2018infant}.

In the experiments, all models (except the underlying model of the official OpenPose library) were trained using a standardized optimization procedure. Pretraining on the general-purpose MPII HPE dataset~\citep{andriluka20142d} was performed, followed by fine-tuning on the training set of In-Motion Poses using the Adam optimizer for 100 epochs with a learning rate of 0.001. We applied data augmentation with random horizontal flipping, scaling ($0.75-1.25$), and rotation ($+$/$-$ 45 degrees). The optimization procedure was obtained through tuning of models on the validation set of In-Motion Poses.

\subsection{Evaluation protocol and performance metrics}
\label{sec:metrics}

\begin{figure}[!t]
\centerline{\includegraphics[width=\columnwidth]{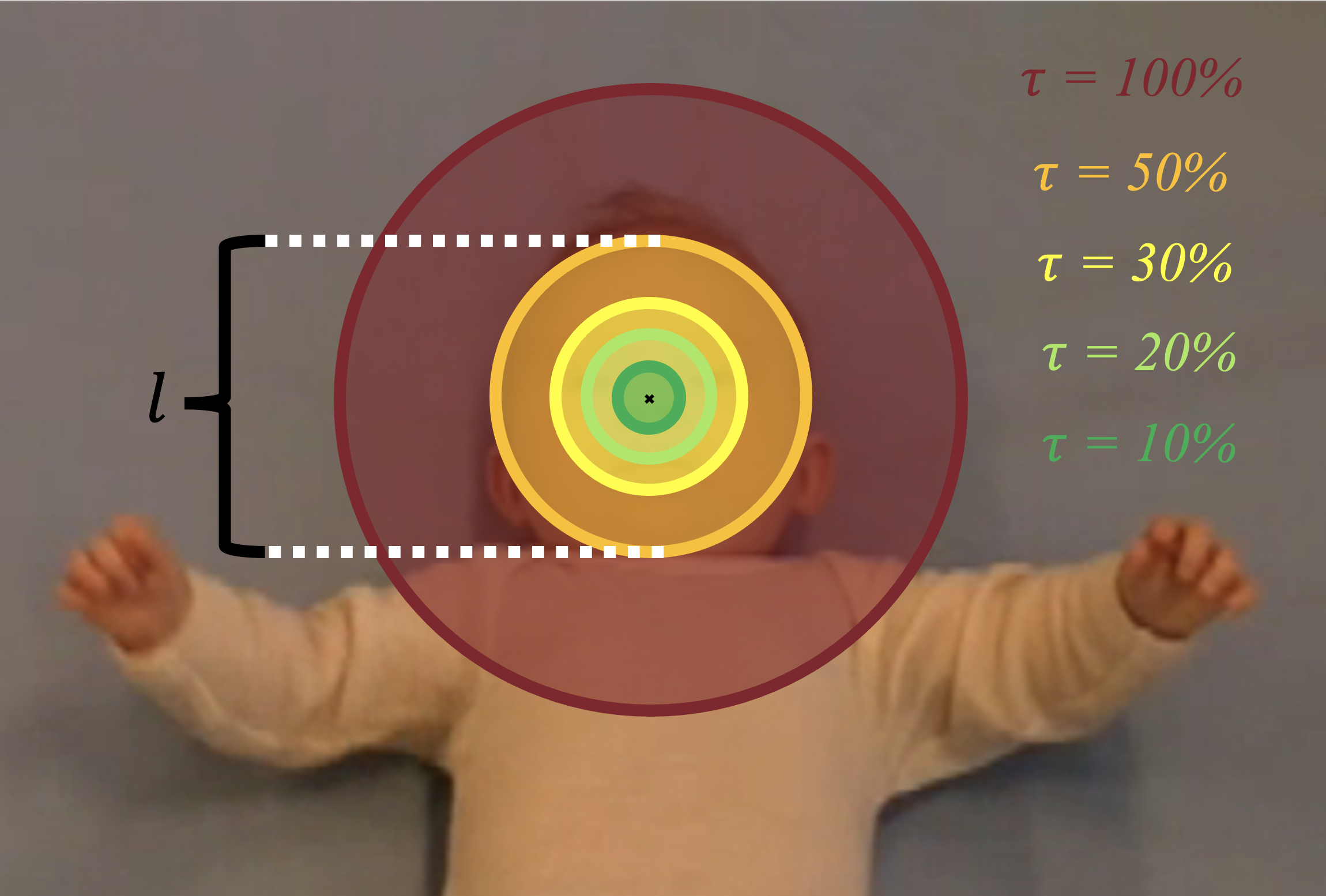}}
\caption{$PCK_{h}@\tau$, the percentage of predictions within $\tau l$ distance from the ground truth location (e.g., nose), is computed across five different thresholds $\tau$ (i.e., 100\%, 50\%, 30\%, 20\%, and 10\%), evaluating the localization performance of a model, from coarse to fine.}
\label{fig:pckh}
\end{figure}

To evaluate the localization performance of the models included in the comparative analysis, positions of body keypoints were predicted on the separate test set of In-Motion Poses, comprising 4 024 images. The retrained OpenPose, CIMA-Pose, EfficientPose, and EfficientHourglass were evaluated using the model outputs upscaled to input resolution with bilinear interpolation (e.g., three transposed convolutions, each with a stride of 2 and 4$\times$4 kernel, performed 8$\times$ upscaling in OpenPose, to increase the spatial resolution of outputs from 46$\times$46 to 368$\times$368), omitting the expensive multi-scale testing and flipping procedure commonly used for benchmarking HPE~\citep{tang2018deeply,yang2017learning}, whereas default post-processing was employed with the official version of OpenPose. Model localization performance was determined by comparing the model outputs to human annotations. The performance metrics included percentage of correct keypoints according to head size ($PCK_{h}@\tau$), normalized mean error ($ME$), and a proposed metric; percentage of correct keypoints according to human-level performance ($PCK_{h}@Human^{0.95}$). $PCK_{h}@\tau$ computes the fraction of keypoints within $\tau l_{i}$ distance from the annotated position, where $l_{i}$ is the infant head length of image $i$. To account for both model robustness and performance in high-precision pose estimation, we calculated measures of $PCK_{h}@\tau$ across various percentages $\tau$ of the head size (see Fig.~\ref{fig:pckh}). Coarse evaluation was performed with $PCK_{h}@1.0$, $PCK_{h}@0.5$, and $PCK_{h}@0.3$, and fine-grained evaluation by $PCK_{h}@0.2$ and $PCK_{h}@0.1$. Moreover, the $ME$ measure reflects the average localization performance of model $m$ on body part $b$ in terms of the mean distance of a model’s predictions to the ground truth locations:

\begin{equation}
ME_{m,b}=\frac{1}{S} \sum_{i=1}^{S} d_{m,b,i}
\label{eq:me}
\end{equation}

where $d_{m,b,i}=\frac{\sqrt{(x_{m,b,i}-\hat{x}_{b,i})^2+(y_{m,b,i}-\hat{y}_{b,i})^2}}{l_{i}}$ is the Euclidean distance from the estimated keypoint position $(x_{m,b,i},y_{m,b,i})$ of model $m$ to the human annotation $(\hat{x}_{b,i},\hat{y}_{b,i})$, for keypoint $b$ in image $i$ of the test set. $ME$ was normalized with respect to the head length $l_{i}$. To compare model performance against human-level performance, we introduce a metric, called $PCK_{h}@Human^{0.95}$. $PCK_{h}@Human^{0.95}$ defines the percentage of model predictions within the 95th percentile of the inter-rater spread of human experts:

\begin{equation}
{PCK_{h}@Human^{0.95}}_{m,b}=\frac{1}{S} \sum_{i=1}^{S} \delta(d_{m,b,i})
\label{eq:pckhuman}
\end{equation}

\begin{equation}
\delta(d_{m,b,i})=\begin{cases}
    1, & \text{if $d_{m,b,i} \leq {H_{b}}^{0.95}$}\\
    0, & \text{otherwise}
  \end{cases}
\label{eq:delta}
\end{equation}

Here, $\delta$ is a binary step function with threshold ${H_{b}}^{0.95}$ defining the 95th percentile of the inter-rater spread (where the mean inter-rater spread ${H_{b}}$ is specified in Equation~\ref{eq:h}). In other words, $PCK_{h}@Human^{0.95}$ is equivalent to $PCK_{h}@\tau$ when ${H_{b}}^{0.95} = \tau$. Thus, $PCK_{h}@Human^{0.95}=95\%$ reflects human-level performance. By utilizing the intraclass correlation coefficient ($ICC$) proposed by~\citet{fisher1992statistical}, we also compared consistency (i.e., $ICC(C,1)$) and agreement (i.e., $ICC(A,1)$) between model localization error and inter-rater spread across body parts. The $ICC$ values, and associated 95\% confidence intervals, between the model $ME$ and the inter-rater spread $H$ of the human experts were calculated using a two-way model. Perfect agreement and consistency with inter-rater spread across body keypoints (i.e., $ICC(A,1)=ICC(C,1)=1$) will suggest that a model displays human-level performance.  

In addition to model localization performance, we evaluated the computational efficiency of the ConvNet models. We provide measures for model complexity (number of parameters), computational cost (FLOPs), and inference time (latency). The inference latency per image was estimated from model predictions on an NVIDIA GTX 1080 Ti GPU with TensorFlow 2.5, CUDA 11.0, and CUDNN 8.1. We used a batch size of 128 and computed the median latency in milliseconds over 10 computational runs.

\subsection{Sample efficiency}
\label{sec:sampleefficiency}

To assess the amount of training data required for ConvNets to converge on the task of infant pose estimation, we carried out experiments with variation in the number of images in the training set, across a range of samples from no fine-tuning\footnote{When models were evaluated without fine-tuning, predictions were made only on the subset of 16 body keypoints that were available both in the MPII dataset and In-Motion Poses.} to 100 images to the full training set of 14 483 infant frames. To evaluate differences in sample efficiency between different ConvNet architectures, experiments were carried out for the most accurate ConvNet in each of the four model families. All experiments were performed over 100 epochs of training, and model performance in $ME$, $PCK_{h}@0.5$, and $PCK_{h}@0.1$ were calculated on the test set of In-Motion Poses. The smaller training samples were constructed by randomly selecting a subset of frames from the original training set, while maintaining the distribution of videos proposed in Section~\ref{sec:inmotionposes}. Hence, the smaller samples and the full training set have similar variation in recording setups.

\section{Results}
\label{sec:results}

Table~\ref{tab:performance} gives an overview of the performances of the eight different ConvNets, as well as the official version of OpenPose, on In-Motion Poses. In terms of localization performance, a 6-37\% decrease in $ME$ compared to the OpenPose baseline is achieved. This is supported by a higher robustness (i.e., gains in $PCK_{h}@1.0$, $PCK_{h}@0.5$, and $PCK_{h}@0.3$). In high-precision pose estimation, $PCK_{h}@0.1$ from 58.71\% to 81.11\% can be observed, compared to 54.89\% and 49.66\% for fine-tuned OpenPose and official OpenPose, respectively. With regards to computational efficiency, all models are smaller, with 1.4-54 times fewer parameters, and require less computation than OpenPose, i.e., 2.2-169 times less FLOPs. Moreover, the most computationally efficient ConvNet, EfficientPose RT, achieved run-time performance of 198 frames per second.

\begin{table*}[]
\footnotesize
\centering
\caption{The performance of the different ConvNets, pretrained on MPII~\citep{andriluka20142d} and fine-tuned on In-Motion Poses, as well as the official OpenPose library~\citep{openpose2021}, in terms of localization performance on the test set of In-Motion Poses, and computational efficiency of the ConvNets from run-time experiments on an NVIDIA GTX 1080 Ti GPU}
\label{tab:performance} 
\begin{tabular}{lllllllllll}
\hline\noalign{\smallskip}
 &  & \multicolumn{6}{c}{\textbf{Localization performance}} & \multicolumn{3}{c}{\textbf{Computational efficiency}} \\ \textbf{Model}
                                & \textbf{Resolution}                                      & $\mathbf{@1.0}^{\mathrm{a}}$ & $\mathbf{@0.5}^{\mathrm{a}}$ & $\mathbf{@0.3}^{\mathrm{a}}$ & $\mathbf{@0.2}^{\mathrm{a}}$ & $\mathbf{@0.1}^{\mathrm{a}}$ & $\mathbf{ME}$       & \textbf{Parameters} & \textbf{FLOPs}  & \textbf{Latency} \\
\noalign{\smallskip}\hline\noalign{\smallskip}
OpenPose library               & -                              & 96.99\%           & 95.51\%          & 90.90\%   & 81.49\%       & 49.66\%          & $0.1432^{\mathrm{b}}$  & -          & - & 62.33$^{\mathrm{c}}$ ms       \\
\noalign{\smallskip}\hline\noalign{\smallskip}
OpenPose               & 368$\times$368                              & 99.94\%           & 99.61\%          & 97.65\%   & 90.40\%       & 54.89\%          & 0.1087  & 26 011 743          & 161 077 013 640 & 35.21 ms        \\
CIMA-Pose              & 368$\times$368                              & 99.98\%           & 99.83\%          & 98.74\%     & 93.09\%     & 59.69\%          & 0.0988 & 2 380 495           & 15 645 092 494  & 11.49 ms        \\
EfficientPose RT       & 224$\times$224                              & 99.96\%           & 99.69\%          & 98.15\%     & 92.15\%     & 58.71\%          & 0.1022 & 481 336             & 955 490 248     & 5.06 ms         \\
EfficientPose I        & 256$\times$256                              & 99.98\%           & 99.83\%          & 98.81\%     & 93.68\%     & 60.78\%          & 0.0974 & 743 476             & 1 785 432 722   & 7.05 ms         \\
EfficientPose II       & 368$\times$368                              & 99.97\%           & 99.84\%          & 98.54\%     & 92.41\%     & 62.25\%          & 0.0969 & 1 759 372           & 7 944 292 598   & 19.38 ms        \\
EfficientPose III      & 480$\times$480                              & 99.99\%           & 99.94\%          & 99.54\%     & 97.57\%     & 78.21\%          & 0.0732 & 3 258 888           & 23 777 830 318  & 41.92 ms        \\
EfficientPose IV       & 600$\times$600                              & 99.98\%           & 99.93\%          & 99.45\%     & 96.77\%     & 71.10\%          & 0.0834 & 6 595 430           & 73 621 311 041  & 96.48 ms       \\
EfficientHourglass B4  & 608$\times$608                              & 99.99\%           & 99.95\%          & 99.56\%     & 97.67\%     & 81.11\%          & 0.0681 & 18 699 936          & 27 009 544 472  & 47.01 ms     \\  
\noalign{\smallskip}\hline
\multicolumn{11}{p{500pt}}{$^{\mathrm{a}}PCK_{h}@1.0$, $PCK_{h}@0.5$, $PCK_{h}@0.3$, $PCK_{h}@0.2$, and $PCK_{h}@0.1$ are abbreviated as $@1.0$, $@0.5$, $@0.3$, $@0.2$, and $@0.1$, respectively.} \\
\multicolumn{11}{p{500pt}}{$^{\mathrm{b}}$Keypoints in certain images, where the OpenPose library lack predictions due to not being confident, are excluded in computation of $ME$.} \\
\multicolumn{11}{p{500pt}}{$^{\mathrm{c}}$Latency estimate of the OpenPose library includes time required to pre-process images and perform default post-processing of ConvNet predictions.}
\end{tabular}
\end{table*}

\begin{figure*}[!t]
\centerline{\includegraphics[width=\textwidth]{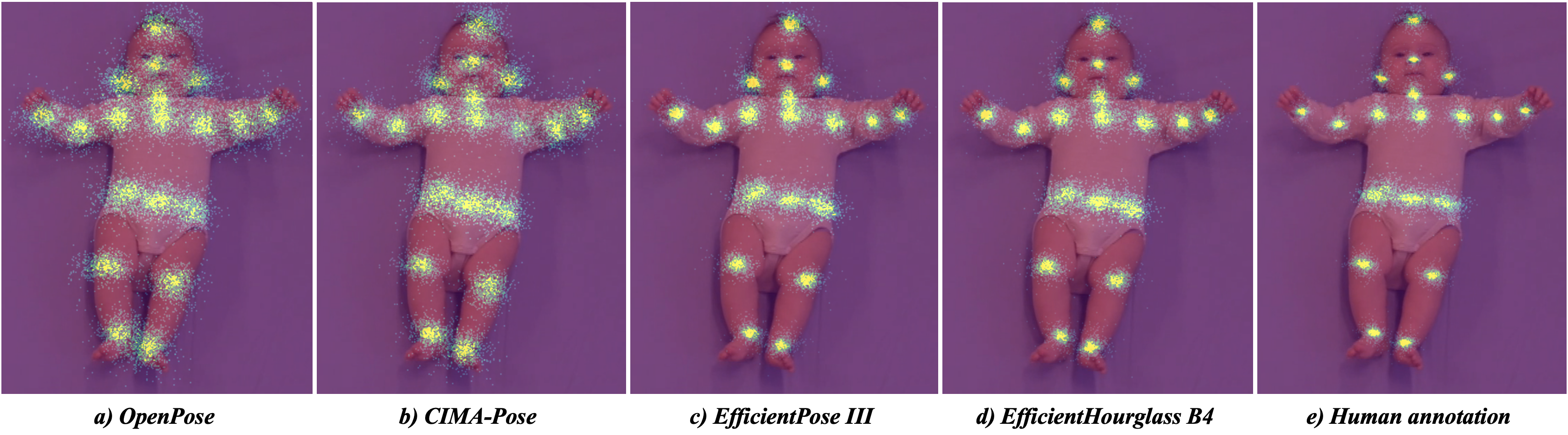}}
\caption{From left: a-d) The distribution of model prediction errors of the different ConvNets on 1 000 randomly sampled frames (according to the distribution of standardized hospital recordings, home-based smartphone recordings, and less standardized hospital recordings) from the test set of In-Motion Poses across body parts, and e) the distribution of the inter-rater spread of the 10 human experts across 100 inter-rater frames (i.e., a total of 1 000 annotations). The prediction errors are normalized according to the head size of the infant in the sample image.}
\label{fig:errors}
\end{figure*}

Table~\ref{tab:humanperformance} displays the localization performance of the top-performing ConvNet of each model family. The most accurate model, EfficientHourglass B4, achieved an $ME$ of 0.0681 compared to the average human inter-rater spread $H$ of 0.0534. This equals an average percentage of human-level performance (i.e., $PCK_{h}@Human^{0.95}$) of 86.71\%, compared to 62.03\% for OpenPose. Fig.~\ref{fig:errors} shows a close resemblance between the spread of the human annotations and the estimates of EfficientPose III and EfficientHourglass B4 across body keypoints. This resemblance was supported by a significant consistency, $ICC(C,1)$, and high agreement, $ICC(A,1)$, between the spread of human expert annotations and the mean error of EfficientPose III and EfficientHourglass B4 (see Table~\ref{tab:icc}). The lower $ICC(A,1)$ compared to $ICC(C,1)$ reflects a slightly higher $ME$ for the ConvNet models compared to the inter-rater spread $H$ of the human experts. A similar resemblance with human annotations was not achieved with OpenPose. 

\begin{table*}[]
\footnotesize
\centering
\caption{The localization performance of OpenPose, CIMA-Pose, EfficientPose III, and EfficientHourglass B4, all pretrained on MPII~\citep{andriluka20142d} and fine-tuned on In-Motion Poses, on the test set of In-Motion Poses, in relation to human-level performance (i.e., inter-rater spread $H$) across body parts $b$, as evaluated by the proposed $PCK_{h}@Human^{0.95}$ metric.}
\label{tab:humanperformance} 
\begin{tabular}{lllllll}
\hline\noalign{\smallskip}
 &  &  & \multicolumn{4}{c}{$\mathbf{PCK_{h}@Human^{0.95}}$} \\
 $\mathbf{b}$ & $\mathbf{H_{b}}$ &    $\mathbf{{H_{b}}^{0.95}}$  & \textbf{OpenPose} & \textbf{CIMA-Pose} & \textbf{EfficientPose III} & \textbf{EfficientHourglass B4} \\
\noalign{\smallskip}\hline\noalign{\smallskip}
Head top                    & 0.0554            & 0.1158             & 60.39\%            & 57.60\%             & 81.59\%                     & 89.31\%                         \\
Nose                        & 0.0301            & 0.0574              & 32.03\%            & 42.89\%             & 74.48\%                     & 82.41\%                         \\
Right ear                   & 0.0603            & 0.1906             & 88.57\%            & 92.40\%             & 94.41\%                   & 92.00\%                         \\
Left ear                    & 0.0502            & 0.1364             & 73.31\%            & 77.49\%             & 88.54\%                     & 89.04\%                         \\
Upper neck                  & 0.0527            & 0.1212             & 80.67\%            & 83.23\%             & 88.77\%                     & 89.19\%                         \\
Right shoulder              & 0.0531            & 0.1106             & 62.97\%            & 73.14\%             & 85.71\%                   & 86.63\%                         \\
Right elbow                 & 0.0429            & 0.0956              & 52.81\%            & 71.00\%             & 81.71\%                     & 86.73\%                         \\
Right wrist                 & 0.0386            & 0.0851              & 45.43\%            & 60.93\%             & 80.14\%                     & 82.60\%                         \\
Upper chest                 & 0.0643            & 0.1200             & 69.38\%            & 72.44\%             & 77.31\%                     & 79.42\%                         \\
Left shoulder               & 0.0576            & 0.1204             & 63.25\%            & 60.71\%             & 88.07\%                     & 88.74\%                         \\
Left elbow                  & 0.0418            & 0.0959              & 48.19\%            & 46.92\%             & 82.50\%                     & 85.69\%                         \\
Left wrist                  & 0.0388            & 0.0901              & 48.83\%            & 52.44\%             & 79.08\%                     & 84.74\%                         \\
Mid pelvis                  & 0.0781            & 0.1587             & 82.75\%            & 82.50\%             & 86.43\%                     & 90.01\%                         \\
Right pelvis                & 0.0812            & 0.1553             & 78.31\%            & 80.89\%             & 87.30\%                     & 88.72\%                         \\
Right knee                  & 0.0549            & 0.1119             & 66.58\%            & 77.24\%             & 86.63\%                     & 89.02\%                         \\
Right ankle                 & 0.0417            & 0.0902              & 51.07\%            & 60.21\%             & 75.47\%                     & 80.79\%                         \\
Left pelvis                 & 0.0828            & 0.1603             & 79.25\%            & 77.53\%             & 88.07\%                     & 90.31\%                         \\
Left knee                   & 0.0489            & 0.1049             & 49.06\%            & 48.29\%             & 88.22\%                     & 89.71\%                         \\
Left ankle                  & 0.0408            & 0.0861              & 45.75\%            & 47.24\%             & 75.70\%                     & 82.38\%                         \\
\textit{All body parts}     & \textit{0.0534}   & \textit{0.1161}    & \textit{62.03\%}   & \textit{66.58\%}    & \textit{81.59\%}            & \textit{86.71\%}    \\
\noalign{\smallskip}\hline
\end{tabular}
\end{table*}

\begin{table*}[]
\footnotesize
\centering
\caption{Absolute agreement and consistency (i.e., $ICC(A,1)$ and $ICC(C,1))$ of ConvNets in relation to human expert inter-rater spread across body parts, with 95\% confidence intervals in brackets}
\label{tab:icc} 
\begin{tabular}{lllll}
\hline\noalign{\smallskip}
         & \textbf{OpenPose} & \textbf{CIMA-Pose} & \textbf{EfficientPose III} & \textbf{EfficientHourglass B4} \\
\noalign{\smallskip}\hline\noalign{\smallskip}
$ICC(A,1)$ & 0.00 [-0.03, 0.07]     & 0.08 [-0.04, 0.32]    & 0.47 [-0.03, 0.84]                & 0.64 [-0.03, 0.91]                     \\
$ICC(C,1)$ & 0.02 [-0.43, 0.46]    & 0.45 [0.01, 0.75]       & 0.94 [0.85, 0.98]                 & 0.96 [0.91, 0.99]   \\
\noalign{\smallskip}\hline
\end{tabular}
\end{table*}

\begin{figure*}[!t]
\centerline{\includegraphics[width=\textwidth]{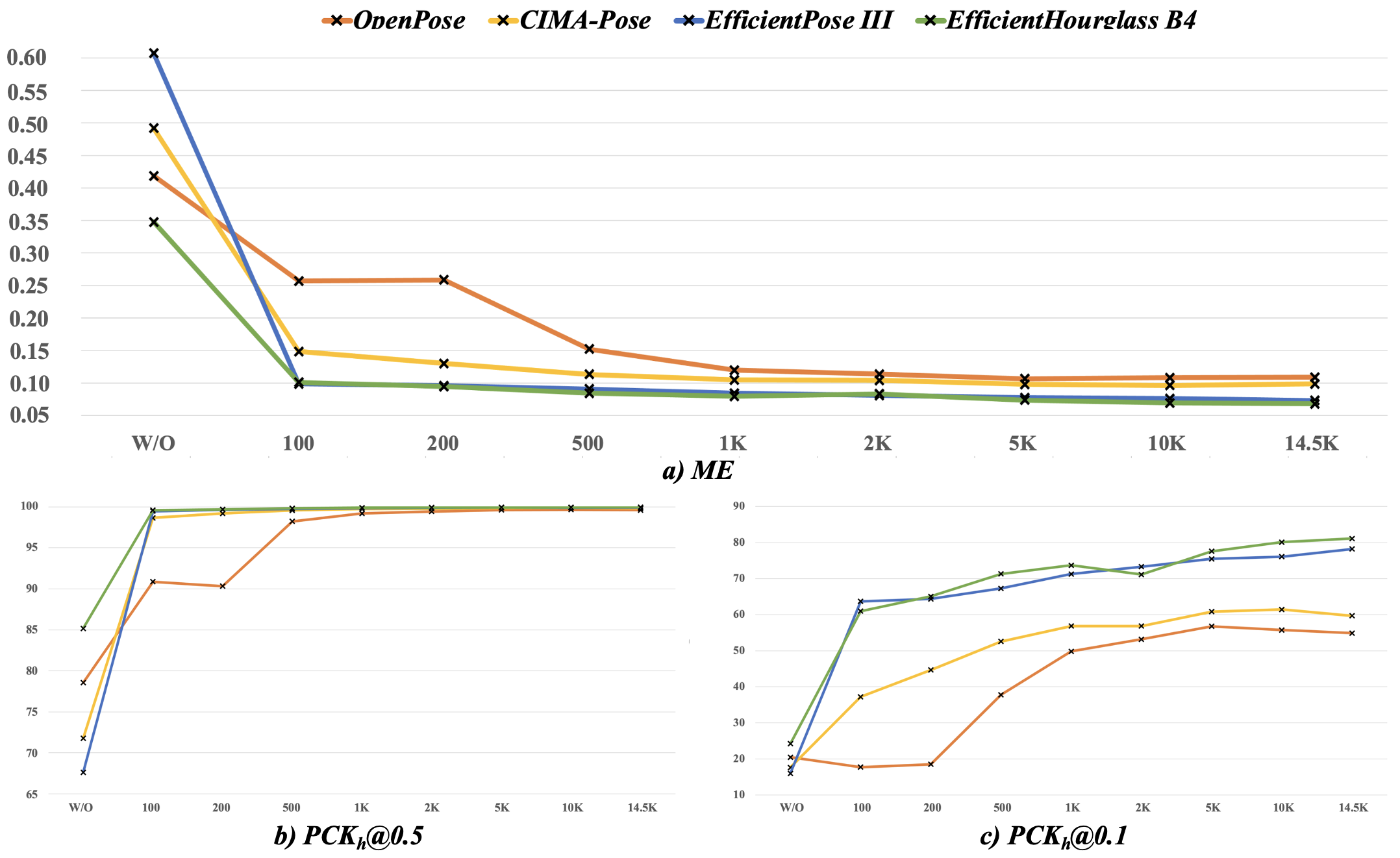}}
\caption{Localization performance of OpenPose, CIMA-Pose, EfficientPose III, and EfficientHourglass B4, all pretrained on MPII~\citep{andriluka20142d}, without fine-tuning (i.e., W/O) and with increasing amounts of data (from 100 to 14 483 images) for fine-tuning on In-Motion Poses.}
\label{fig:samples}
\end{figure*}

Fig.~\ref{fig:samples} illustrates that fine-tuning significantly improves localization performance of infant pose estimation compared to no fine-tuning (i.e., W/O). Moreover, all ConvNets benefit from increased training set size, especially in terms of the $PCK_{h}@0.1$ measure (Fig.~\ref{fig:samples}c). However, whereas localization performance of OpenPose and CIMA-Pose saturates at sample sizes beyond 5 000 images, EfficientPose III and EfficientHourglass B4 benefit from larger training sets. There is also a tendency that EfficientPose III and EfficientHourglass are more stable across dataset sizes, with a smaller difference in localization performance from 100 to 14 483 images, compared to OpenPose and CIMA-Pose.

\begin{figure*}[!t]
\centerline{\includegraphics[width=\textwidth]{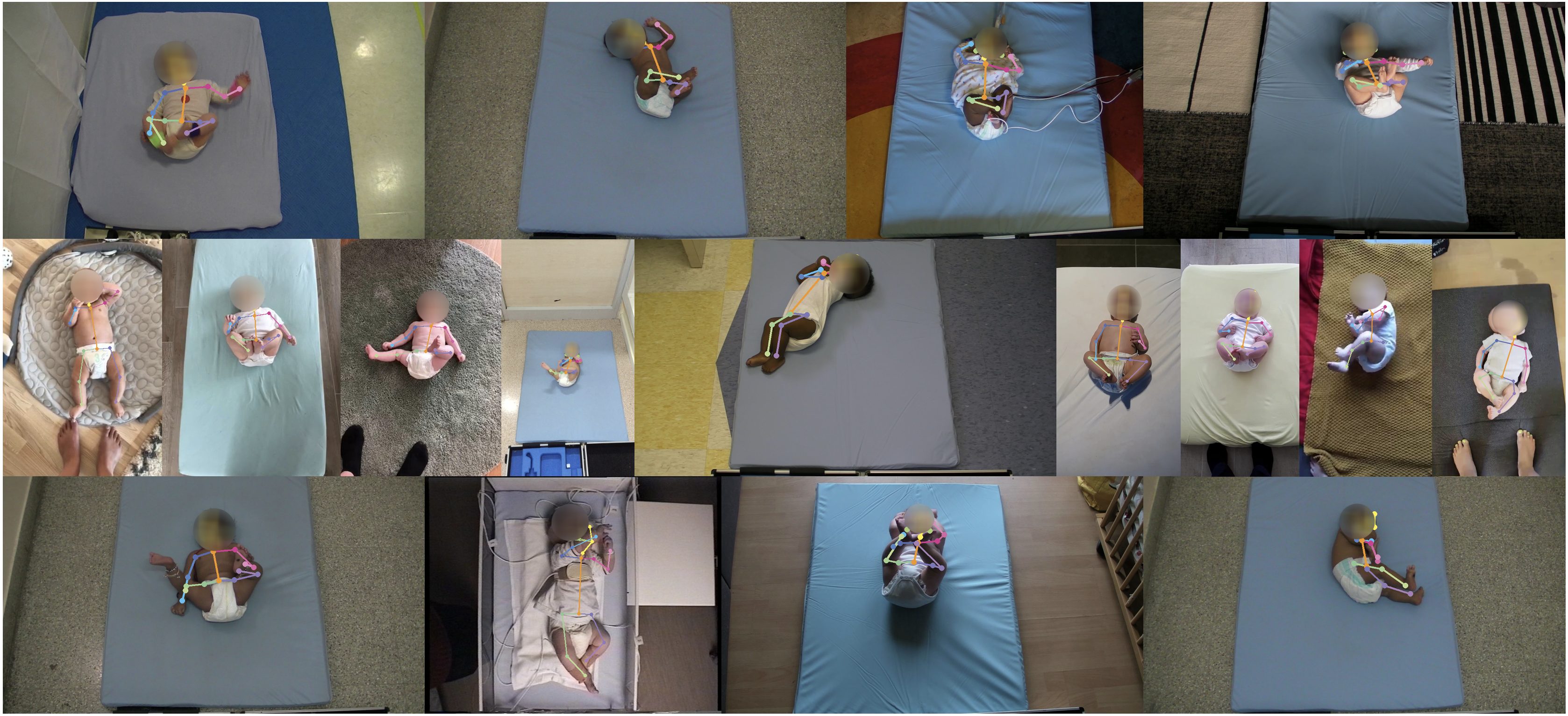}}
\caption{Predictions of EfficientHourglass B4 on rare but normal infant poses in the test set of In-Motion Poses. The first and second row contain images where the model correctly predicted the position of body keypoints. The third row indicates cases where the model missed certain body keypoints (images from left to right: 1) right ankle, 2) head top and nose, 3) right elbow and right wrist, and 4) right wrist and left wrist). Infant faces are blurred to ensure anonymity.}
\label{fig:challenging}
\end{figure*}

In Fig.~\ref{fig:challenging}, the localization performance of EfficientHourglass B4 is assessed qualitatively by providing model predictions on a selection of challenging images (i.e., less frequently occurring infant poses as described in Section~\ref{sec:inmotionposes}) in the test set of In-Motion Poses.

\newpage
\section{Discussion}
\label{sec:discussion}

The main objective of the study was to obtain computationally efficient markerless infant pose estimation with a level of localization performance approaching that of human expert annotations. A comparative analysis has showed that performance levels comparable to human expert performance can be achieved, by utilizing contemporary ConvNets for HPE together with an extensive infant video database. This is reflected by $PCK_{h}@Human^{0.95}$ of the top-performing ConvNets approaching human-level performance, whereas the commonly applied OpenPose network does not reach similar level of localization performance.

\subsection{Improving localization performance}
\label{sec:localizationperformance}

The large improvement in localization performance compared to the state-of-the-art method OpenPose~\citep{cao2018openpose} is due to two main reasons. First, the hypothesis of~\citet{sciortino2017estimation}, that HPE ConvNets require fine-tuning on a selection of infant images to perform well on pose estimation of infants, is confirmed. The introduction of a large-scale infant pose dataset, In-Motion Poses, has improved the localization performance of OpenPose from 78.56\% to 99.61\% on $PCK_{h}@0.5$, as illustrated by Fig.~\ref{fig:samples}b. Taking into account the error taxonomy of~\citet{ruggero2017benchmarking}, this indicates that the coarse localization errors, like the frequency of inversions (i.e., the predictions that appear at an incorrect body keypoint, such as misinterpretation of the left and right wrist) and misses (i.e., the erroneous localizations that are made without interfering with other keypoints), have been reduced. Despite the increased robustness with regards to coarse prediction errors, the optimal level of localization performance has not been reached. Further improvement of the ConvNets may be achieved by more systematically studying the cases where the models fall short, for example with substantial occlusion of body parts or specific body postures. Fig.~\ref{fig:challenging} indicates that such scenarios exist. Accordingly, we could extend the existing dataset with images that target these situations to further improve model robustness through retraining. In a future perspective, it would also be valuable to assess if we could take into account the temporal information of a video to reduce prediction errors due to occlusion or rare body postures. Pose tracking that extends beyond frame-by-frame pose estimation may achieve this, but current progress in the field is restricted to processing a single pair of video frames with limited gap in time~\citep{bertasius2019learning}, which may not address cases of prolonged occlusion.

Second, the large improvement in $PCK_{h}@0.2$, $PCK_{h}@0.1$, and $PCK_{h}@Human^{0.95}$ of CIMA-Pose, EfficientPose, and EfficientHourglass B4, compared to OpenPose, is due to a reduction in fine prediction errors. EfficientPose III, EfficientPose IV\footnote{EfficientPose IV displayed lower localization performance than EfficientPose III on In-Motion Poses, due to small batch size during training, which was necessary for the model to fit into GPU memory. As demonstrated by Table~\ref{tab:performance} and Table~\ref{tab:batchsize}, EfficientPose IV performed better than EfficientPose III in case of similar batch sizes.}, and EfficientHourglass B4 reduce fine prediction errors better than OpenPose by operating on increased input and output resolutions. The consistent high resolution of EfficientHourglass B4 seems to maximize this benefit by displaying the highest values of $PCK_{h}@0.1$ and $PCK_{h}@Human^{0.95}$. However, the increase of resolution comes at the cost of reduced computational efficiency, in terms of increased number of FLOPs and decreased latency (see Table~\ref{tab:performance}). Thus, alternative methods for post-processing of ConvNet predictions (e.g., soft-argmax~\citep{levine2016end}), or post-processing of the frame-by-frame position estimates over consecutive frames by low-pass filters, such as median filtering~\citep{tukey1977exploratory}, might reduce fine prediction errors more effectively. However, this demands that the video has a sufficient sample rate (e.g., 60 fps). Furthermore, fine prediction errors may also be minimized by decreasing the spread in annotated keypoint positions. As illustrated in Fig.~\ref{fig:errors}, the distributions of prediction errors of EfficientPose III and EfficientHourglass B4 across body parts resemble the inter-rater spread of the human experts (e.g., higher variation in the placement of the keypoints of the pelvis, compared to the nose keypoint). This indicates that contemporary ConvNets for HPE, when supplied with sufficient amounts of training data (see Fig.~\ref{fig:samples} for the effect of sample size), are able to maximize the benefit of human annotations. Hence, a hypothesis for further studies is that more precisely annotated keypoints will further eliminate fine prediction errors, by model error being highly correlated with the inter-rater spread of human experts (see Table~\ref{tab:icc}). Consequently, lower variation in the annotation of the keypoints of the pelvis may improve the ability of the ConvNets to localize these keypoints with high localization performance. More consistent annotations between human experts, reflected by lower inter-rater spread, may be obtained by proposing more precise definitions of the keypoint positions, than those in Appendix~\ref{sec:keypoints}. This could be particularly valuable for body keypoints that currently have higher inter-rater spread (e.g., for the keypoint of the upper chest). Human expert annotations may also be supplemented or replaced by other methods, such as marker-based solutions and 3D motion capture systems. These approaches may also yield performance improvements beyond fine prediction errors, by providing more precise annotations of occluded keypoints than can be achieved with 2D videos. We suggest that studies on infant pose estimation, and HPE in general (e.g., on challenges such as MPII~\citep{andriluka20142d}), judge localization performance against metrics related to human-level performance, such as $PCK_{h}@Human^{0.95}$, to evaluate the progress on these tasks in relation to human-level performance. 

\subsection{Improving computational efficiency}
\label{sec:efficiency}

Our comparative analysis has shown that a large model size (i.e., number of parameters) is not necessary for high-precision infant pose estimation. On similar input resolution, both OpenPose and CIMA-Pose were outperformed by the more computationally efficient low-complexity EfficientPose II model on $PCK_{h}@0.1$ (see Table~\ref{tab:performance}). Instead, it appears that high-precision infant pose estimation can be obtained with a relatively small number of parameters. This is demonstrated by EfficientPose III displaying only 5.12\% decrease in $PCK_{h}@Human^{0.95}$, compared to EfficientHourglass B4, despite having 5.7 times fewer parameters. Combining this observation with the influence of high input and output resolution on localization performance, we would suggest further studies to investigate the effect of high resolution with low-complexity ConvNets. This could potentially narrow the current gap in localization performance between computationally efficient ConvNets, such as EfficientPose RT, and high-precision counterparts that are less computationally efficient, like EfficientPose III and EfficientHourglass B4. It would also be of particular interest to systematically study the optimal trade-off between localization performance and computational efficiency, by carefully assessing the localization performance of ConvNets of various complexities across different image resolutions. Our study suggests that ConvNets developed for HPE can be simplified when transferred to the infant pose estimation domain. HPE targets more complex circumstances and environments (e.g., images of multiple persons, a wide range of different activities, individuals of varying age, and substantial occlusion), whereas infant pose estimation is concerned with a single, clearly visible infant in supine position according to the guidelines of GMA~\citep{ferrari2004prechtl,andriluka20142d}. Potential paths for reducing network complexity could be 1) a decrease in network width (i.e., number of feature maps), and 2) less extensive use of multi-scale ConvNet architectures. The former may more appropriately address the little diversity in infant videos compared to the far-reaching HPE task, whereas the latter takes into account the small variation in an infant’s distance to the camera and anatomical proportions. Nevertheless, from studying the inference latency of the ConvNets, we observed processing speeds from 10 to 198 fps (Table~\ref{tab:performance}) on an NVIDIA GTX 1080 Ti consumer GPU. Further speedups of the pool of models studied in this paper may be obtained by implementing the ConvNets in low-level code like C++ or CUDA. Thus, a three-minute video of infant spontaneous movements could potentially be processed by a high-precision pose estimator in less than three minutes, which is feasible for clinical use. Moreover, the efficiency of the ConvNets can be further enhanced by utilizing techniques for compressing models with minimal loss of localization performance. Quantization-aware training, knowledge distillation, model pruning, and sparse kernels are paths that are worth to investigate ~\citep{tensorflow2020,bucilua2006model,tung2018clip,elsen2020fast}. By obtaining accelerated and compressed ConvNets, the automatic pose estimation have the potential to be deployed locally at smartphones in the clinic and at home. Thus, infant pose estimation will be more easily applicable, while preserving patient privacy through decentralized processing of infant recordings on local devices.

\subsection{External validity}
\label{sec:validity}

In previous studies on ConvNet-based markerless infant pose estimation from 2D videos, investigations have been restricted to small or synthetic samples of infant videos~\citep{hesse2018computer,chambers2020computer}. Hence, the external validity of such approaches is debatable, since ConvNets require large amounts of realistic images across various settings related to the task at hand to perform well on pose estimation. In this study, we have utilized a large-scale international database of GMA certified video recordings to train the ConvNets. Subsequently, we have validated the models on a separate set of 284 infant videos from a diverse range of hospital and home-based setups (see Fig.~\ref{fig:inmotionposes}a). The high resistance to coarse prediction errors of the evaluated ConvNets suggests that infant pose estimation promotes flexibility in application in real-world scenarios. This encompasses various settings (e.g., clinic, research center, and home), across different countries, and without depending on specific camera equipment. When assessing the transfer validity of the ConvNets fine-tuned on In-Motion Poses on the synthetic dataset proposed by~\citet{hesse2018computer}, only the best performing ConvNet on In-Motion Poses, EfficientHourglass B4, outperformed the official version of the state-of-the-art method OpenPose and displayed an acceptable transfer by maintaining a high level of localization performance (Table~\ref{tab:minirgbd} and Fig.~\ref{fig:transfer}). This could suggest that the high-capacity multi-scale feature extractor of EfficientHourglass B4, through pretraining on MPII~\citep{andriluka20142d} and fine-tuning on In-Motion Poses, has learnt features that generalize beyond the natural infant images of In-Motion Poses. On the contrary, the feature extractors of OpenPose, CIMA-Pose, and EfficientPose are of lower relative capacity and contain fewer abstraction levels (i.e., scales) compared to EfficientHourglass B4 (Fig.~\ref{fig:models}). Hence, these fine-tuned ConvNets might lack the ability for appropriate transfer beyond recording setups of In-Motion Poses (e.g., plain backgrounds, and natural lighting and shading). However, the consistent localization performance of the official OpenPose library~\citep{openpose2021} (Table~\ref{tab:performance} and Table~\ref{tab:minirgbd}) suggests that training on a sufficiently heterogeneous and large-scale human pose dataset, such as COCO~\citep{lin2014microsoft} of 250 000 human poses from various contexts, may combat the lack of high-capacity and multi-scale feature extraction to yield better generalizability. Similar effects could be achieved by combining In-Motion Poses with synthetic or natural infant pose datasets covering the variation in recording setups we want ConvNets to be tuned towards. Nevertheless, we should take into consideration the overall model capacity (i.e., number of parameters), which for CIMA-Pose and EfficientPose might not be sufficient to achieve appropriate transfer from In-Motion Poses to synthetic infants. We could therefore investigate ConvNet compound scaling on infant pose estimation, to determine the appropriate scaling factors of input resolution, network width, and network depth. Further studies should also more thoroughly assess the external validity of the trained ConvNets on real-life infant recordings, to verify that the high level of localization performance demonstrated by the present study indeed can be reproduced. This involves assessing the robustness in operating on video recordings from different recording setups with large variations in aspects, such as video quality, background environment, camera angle, and lighting conditions. The infant pose estimators could also be validated across groups of infants with different age, size, skin color, clothing, and postural variability within datasets like In-Motion Poses. Moreover, the degree of localization performance of the ConvNets in relation to state-of-the-art marker-based motion capture systems could also be assessed~\citep{vicon2020,qualisys2020}. It is worth stressing that it is unrealistic to expect flawless pose estimation in recording situations highly dissimilar to the settings the models have been trained and evaluated in. However, the models can be retrained on other video databases when keypoint annotations are available. It is also worth investigating if the predefined set of body keypoints is sufficient for performing relevant assessments of characteristics of infant spontaneous movements identified in clinical GMA. However, for applications emphasizing movement kinematics of other body keypoints (e.g., rotation of hands and feet, and relative movements of fingers or toes), the proposed infant pose estimation can be extended through retraining of ConvNets on different annotated sets of keypoints. 

In summary, with improved ConvNet architectures and an extensive database of infant video recordings, body keypoint positions can be estimated with human-level performance. This will enable capturing more subtle infant movements and postures, and, consequently, improve early detection of risk-related infant movement kinematics~\citep{ihlen2020machine,einspieler2019cerebral}. These improved ConvNets will also facilitate the assessments of infant movement kinematics which require a high level of detail, like fidgety movements or postural patterns in specific parts of the body, such as side-to-side head movements and atypical head centering~\citep{einspieler2019cerebral}.

\newpage
\section{Conclusions}
\label{sec:conclusion}

The present study represents a significant progress towards clinically feasible markerless pose estimation of infant movements between three to five months of post-term age. This has been achieved by combining state-of-the-art ConvNets for human pose estimation with a novel heterogenous infant dataset. Highly precise detection of body keypoints enables accurate localization of segments and joints, which may facilitate computer-based assessment of characteristics of infant spontaneous movements related to risk of developmental disorders. With no dependency to body-worn markers, sensors or other expensive laboratory equipment, the automatic infant pose estimation can handle videos both captured by parents at home and by physicians at a hospital clinic. In conclusion, this technology has the potential to facilitate further research initiatives on infant movement analysis and motivate national and worldwide collaborations. 

\appendix

\section{Keypoint definitions}
\label{sec:keypoints}

\begin{figure}[!t]
\centerline{\includegraphics[width=\columnwidth]{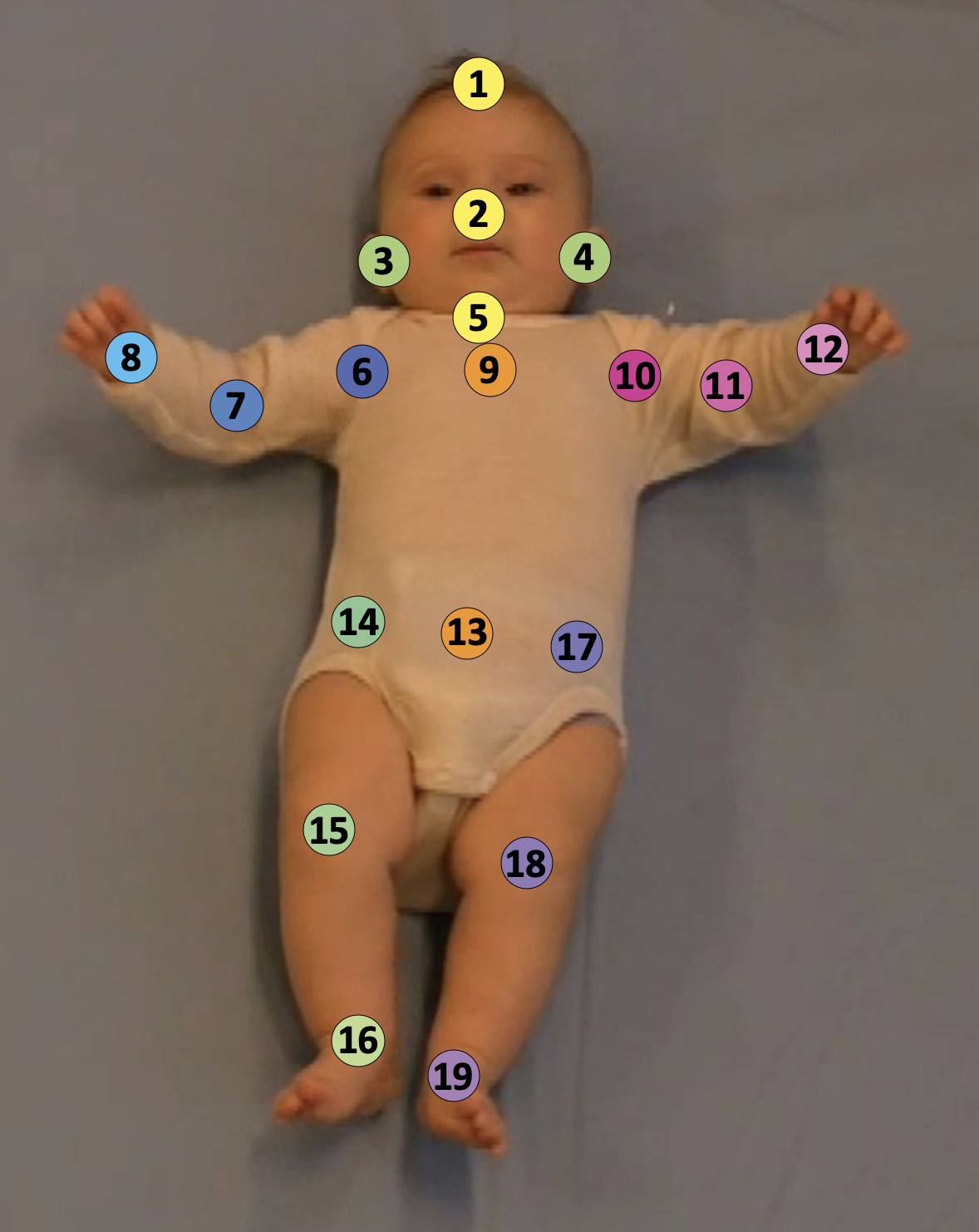}}
\caption{The placements of the 19 different body keypoints on an infant.}
\label{fig:keypoints}
\end{figure}

The set of 19 body keypoints along with their definitions (see Fig.~\ref{fig:keypoints} and Table~\ref{tab:keypoints}) were agreed upon by an expert group of human movement scientists and infant physiotherapists. The body keypoints were selected to cover most effectively the many degrees of freedom in the infant movements, while at the same time being properly defined to facilitate consistent annotation across humans. 

\begin{table}[]
\centering
\caption{Definitions of body keypoints}
\label{tab:keypoints} 
\begin{tabular}{lll}
\hline\noalign{\smallskip}
\textbf{\#} & \textbf{Body keypoint} & \textbf{Definition}                                        \\
\noalign{\smallskip}\hline\noalign{\smallskip}
1           & Head top               & Top of the forehead                                        \\
2           & Nose                   & Tip of the nose                                            \\
3           & Right ear              & Center of the right ear                                    \\
4           & Left ear               & Center of the left ear                                     \\
5           & Upper neck             & Center of the larynx                                       \\
6           & Right shoulder         & Center of the right shoulder joint                         \\
7           & Right elbow            & Center of the right elbow joint                            \\
8           & Right wrist            & Center of the right wrist joint                            \\
9           & Upper chest            & Midway between 6 and 10 \\
10          & Left shoulder          & Center of the left shoulder joint                          \\
11          & Left elbow             & Center of the left elbow joint                             \\
12          & Left wrist             & Center of the left wrist joint                             \\
13          & Mid pelvis             & Midway between 14 and 17                   \\
14          & Right pelvis           & Right spina iliaca anterior superior                       \\
15          & Right knee             & Center of the right knee joint                             \\
16          & Right ankle            & Center of the right ankle joint                            \\
17          & Left pelvis            & Left spina iliaca anterior superior                        \\
18          & Left knee              & Center of the left knee joint                              \\
19          & Left ankle             & Center of the left ankle joint    \\          \noalign{\smallskip}\hline           
\end{tabular}
\end{table}

\section{Batch size inspection}
\label{sec:batchsize}

We assessed the effect of fine-tuning the EfficientPose models on a reduced batch size of four images (i.e., the batch size of EfficientPose IV) to investigate possible performance degrade with EfficientPose IV due to inappropriate batch size. In comparison to Table~\ref{tab:performance}, Table~\ref{tab:batchsize} displays performance degrade from training with reduced batch size, most evident in terms of high-precision localization, with 11.20\% to 30.83\% reduction in $PCK_{h}@0.1$.

\begin{table*}[]
\footnotesize
\centering
\caption{The localization performance of EfficientPose RT and I-III on the test set of In-Motion Poses, when trained with the batch size of EfficientPose IV, followed by the performance difference in relation to the experiments in Table~\ref{tab:performance}}
\label{tab:batchsize} 
\begin{tabular}{lllllll}
\hline\noalign{\smallskip}
 \textbf{Model}
                                                                      & $\mathbf{PCK_{h}@1.0}$ & $\mathbf{PCK_{h}@0.5}$ & $\mathbf{PCK_{h}@0.3}$ & $\mathbf{PCK_{h}@0.2}$ & $\mathbf{PCK_{h}@0.1}$ & $\mathbf{ME}$       \\
\noalign{\smallskip}\hline\noalign{\smallskip}
EfficientPose RT                                    & 99.80\% (-0.16\%)           & 99.32\% (-0.37\%)          & 92.93\% (-5.22\%)      & 72.50\% (-19.65\%)     & 27.88\% (-30.83\%)          & 0.1717 (0.0695)         \\
EfficientPose I                          & 99.94\% (-0.04\%)            & 99.66\% (-0.17\%)          & 97.22\% (-1.59\%)     & 85.42\% (-8.26\%)     & 38.38\% (-22.40\%)          & 0.1311 (0.0336)  \\
EfficientPose II                                 & 99.98\% (0.01\%)           & 99.78\% (-0.06\%)          & 98.01\% (-0.53\%)     & 89.85\% (-2.56\%)     & 49.73\% (-12.52\%)          & 0.1137 (0.0168)   \\
EfficientPose III          & 99.99\% (0.00\%)           & 99.94\% (0.00\%)          & 99.47\% (-0.07\%)     & 96.48\% (-1.09\%)     & 67.01\%  (-11.20\%)         & 0.0884 (0.0152)  \\
\noalign{\smallskip}\hline
\end{tabular}
\end{table*}

\section{Transfer validity}
\label{sec:transfer}

To investigate the transfer validity of the methods in our comparative analysis, we evaluated the localization performance of the models fine-tuned on In-Motion Poses, as well as the official OpenPose library, on the openly available MINI-RGBD dataset proposed by~\citet{hesse2018computer} (Table~\ref{tab:minirgbd}). The MINI-RGBD dataset comprises 12 synthetic infant video recordings of quite different nature than the recordings in In-Motion Poses. Localization performance, in terms of $PCK_{h}@1.0$, $PCK_{h}@0.5$, $PCK_{h}@0.3$, $PCK_{h}@0.2$, $PCK_{h}@0.1$, and $ME$, was measured on the subset of 12 body keypoints that are similar for MINI-RGBD and In-Motion Poses (i.e., nose, upper neck, shoulders, elbows, wrists, knees, and ankles). Since MINI-RGBD does not contain a keypoint for the top of the forehead, the head length of an infant was estimated as two times\footnote{The head length of an infant (i.e., the distance from head top to upper neck) in In-Motion Poses was in average 1.98 times the distance from nose to upper neck.} the distance between the annotated keypoints of the nose and upper neck. This ensures that the evaluation metrics reflect a similar level of correctness as the metrics used with the evaluation on In-Motion Poses in Table~\ref{tab:performance}.

\begin{figure*}[!t]
\centerline{\includegraphics[width=\textwidth]{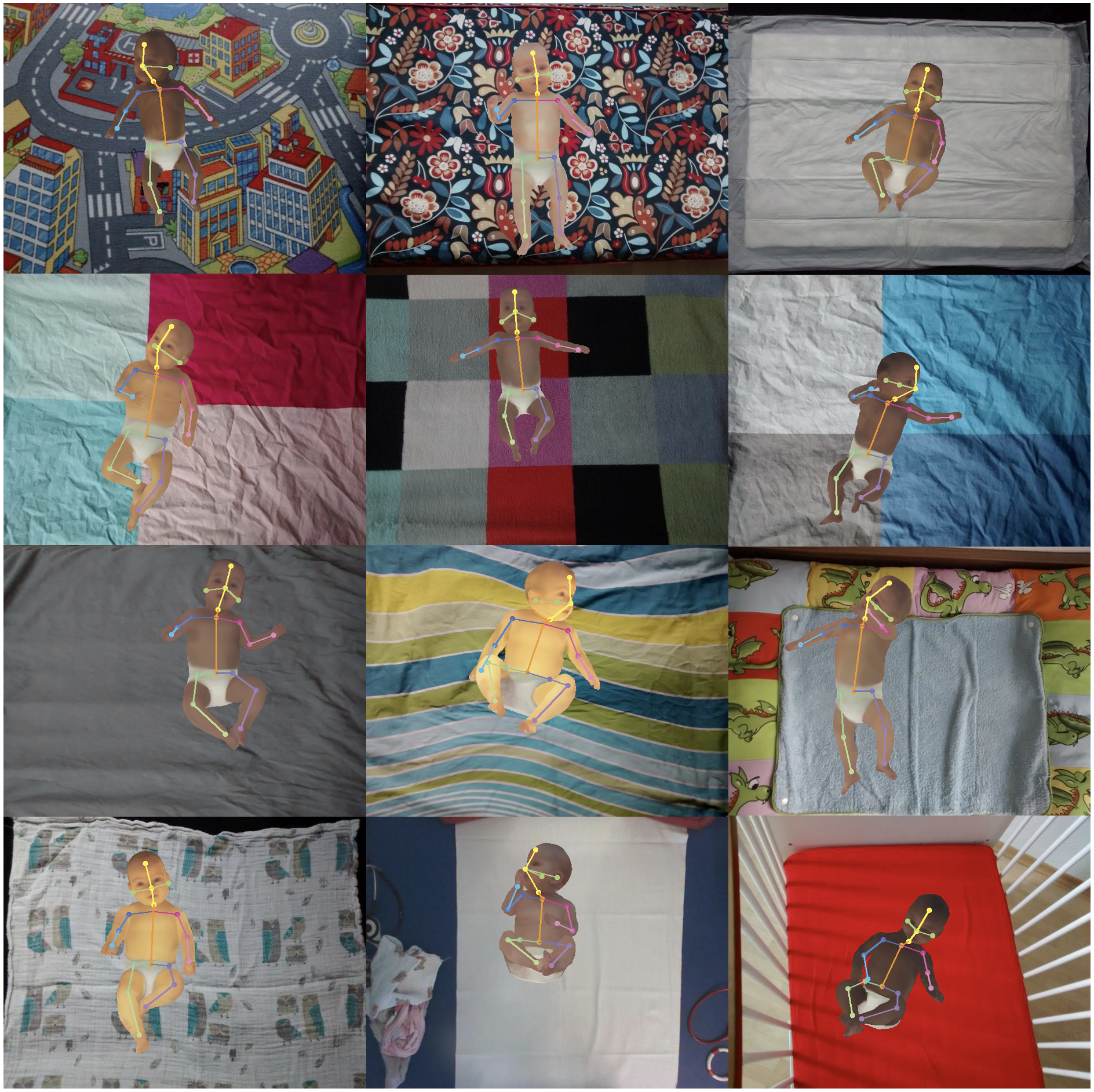}}
\caption{Predictions of keypoint locations of EfficientHourglass B4 in randomly selected frames from videos in the MINI-RGBD dataset~\citep{hesse2018computer}.}
\label{fig:transfer}
\end{figure*}

Furthermore, for the most accurate ConvNet, namely EfficientHourglass B4, we conducted a qualitative experiment by estimating the locations of the 19 body keypoints in In-Motion Poses on a randomly selected frame in each of the 12 infant videos in the MINI-RGBD dataset (Fig.~\ref{fig:transfer}).

We also supply as supplementary material frame-by-frame predictions of keypoint locations in a real, external infant recording for the best performing ConvNet in each model family, as well as by the use of the official version of OpenPose. The recording follows the standards for GMA~\citep{einspieler2005prechtl}, and has been recorded using the setup of the In-Motion App~\citep{adde2021motion}, which is similar to the home-based smartphone recordings in In-Motion Poses.

\begin{table*}[]
\footnotesize
\centering
\caption{The transfer validity of the different ConvNets, pretrained on MPII~\citep{andriluka20142d} and fine-tuned on In-Motion Poses, and the official OpenPose library~\citep{openpose2021}, in terms of localization performance on the MINI-RGBD dataset}
\label{tab:minirgbd} 
\begin{tabular}{lllllll}
\hline\noalign{\smallskip}
 \textbf{Model}
                                                                      & $\mathbf{PCK_{h}@1.0}$ & $\mathbf{PCK_{h}@0.5}$ & $\mathbf{PCK_{h}@0.3}$ & $\mathbf{PCK_{h}@0.2}$ & $\mathbf{PCK_{h}@0.1}$ & $\mathbf{ME}$       \\
\noalign{\smallskip}\hline\noalign{\smallskip}
OpenPose library                  & 98.35\%           & 97.02\%          & 94.47\%     & 90.75\%     & 73.80\%          & 0.1030         \\
\noalign{\smallskip}\hline\noalign{\smallskip}
OpenPose                          & 88.59\%           & 79.59\%          & 71.77\%     & 62.27\%     & 38.41\%          & 0.3926         \\
CIMA-Pose                         & 95.72\%           & 88.99\%          & 81.27\%     & 71.83\%     & 46.68\%          & 0.2415         \\
EfficientPose RT                  & 94.98\%           & 91.28\%          & 86.91\%     & 79.98\%     & 53.83\%          & 0.2135         \\
EfficientPose I                   & 93.13\%           & 91.09\%          & 88.16\%     & 81.98\%     & 56.19\%          & 0.2772         \\
EfficientPose II                  & 92.49\%           & 90.41\%          & 87.41\%     & 80.57\%     & 54.60\%          & 0.3263         \\
EfficientPose III                 & 83.79\%           & 81.45\%          & 79.60\%     & 76.06\%     & 58.56\%          & 0.8559         \\
EfficientPose IV                  & 93.02\%           & 91.15\%          & 89.05\%     & 86.14\%     & 71.35\%          & 0.2565         \\
EfficientHourglass B4             & 99.81\%           & 99.17\%          & 97.52\%     & 94.13\%     & 75.86\%          & 0.0845         \\
\noalign{\smallskip}\hline
\end{tabular}
\end{table*}

\section*{Acknowledgments}
\label{sec:acknowledgment}

This study was possible only due to the unified In-Motion research initiative on computer-based assessment of infant spontaneous movements and prediction of cerebral palsy, resulting in the multi-site database of infant recordings. The authors would like to acknowledge the following key personnel and institutions contributing in collecting video recordings: Norway; Toril Larsson Fjørtoft at St. Olavs University Hospital, Inger Elisabeth Silberg at Oslo University Hospital, Nils Thomas Songstad at University Hospital of North Norway, Angelique Tiarks at Levanger Hospital, Henriette Paulsen at Vestfold Hospital Trust, India; Niranjan Thomas at Christian Medical College Vellore, United States; Colleen Peyton at University of Chicago Comer Children’s Hospital, Raye-Ann de Regnier and Lynn Boswell at Ann \& Robert H Lurie Children's Hospital of Chicago, Turkey; Akmer Mutlu at Hacettepe University, Belgium; Aurelie Pascal at Ghent University, Denmark; Annemette Brown at Nordsjællands Hospital Hillerød, Great Britain; Anna Basu at Newcastle upon Tyne Hospitals. This work was supported by the Liaison Committee between the Central Norway Regional Health Authority and the Norwegian University of Science and Technology under project number 90056100, the Joint Research Committee between St. Olavs University Hospital and the Faculty of Medicine and Health Sciences, Norwegian University of Science and Technology, the DeepInMotion project funded by the Research Council of Norway with grant number 327146, and RSO Funds from the Faculty of Medicine and Health Sciences, Norwegian University of Science and Technology under project number 81115200.


\bibliographystyle{cas-model2-names}

\bibliography{refs}

\end{document}